\documentclass[sigconf]{acmart}

\renewcommand\footnotetextcopyrightpermission[1]{} 

\usepackage{booktabs} 
\usepackage[utf8]{inputenc} 
\usepackage[T1]{fontenc}    
\usepackage{hyperref}       
\usepackage{url}            
\usepackage{booktabs}       
\usepackage{amsfonts}       
\usepackage{nicefrac}       
\usepackage{microtype}      
\usepackage{hhline}
\usepackage{tikz}
\usetikzlibrary{bayesnet}
\usepackage{graphicx}
\usepackage{amsmath}
\usepackage{multirow}
\usepackage{bbm}
\usepackage{caption,subcaption}
\usepackage{enumitem} 
\usepackage{algorithm}
\usepackage{algorithmicx}
\usepackage{algpseudocode}
\usepackage[colorinlistoftodos]{todonotes}

\usepackage{caption}
\usepackage{caption} 
\setcopyright{rightsretained}
\acmConference[KDD'18]{ACM conference}{August 2018}{London, United Kingdom}
\acmYear{2018}
\copyrightyear{2018}
\acmArticle{4}
\acmPrice{15.00}

\begin{document}
\title{Distinguishing Question Subjectivity from Difficulty for Improved Crowdsourcing}


\author{Yuan Jin, Ye Zhu}
\affiliation{%
  \institution{School of Information Technology\\Deakin University, Australia}
}
\email{{yuan.jin, ye.zhu}@deakin.edu.au}

\author{Mark Carman, Wray Buntine}
\affiliation{%
  \institution{Faculty of Information Technology\\Monash University, Australia}
}
\email{{mark.carman, wray.buntine}@monash.edu}

\begin{abstract}
The questions in a crowdsourcing task typically exhibit varying degrees of \textit{difficulty} and \textit{subjectivity}. Their joint effects give rise to the variation in responses to the same question by different crowd-workers. This variation is low when the question is easy to answer and objective, and high when it is difficult and subjective. Unfortunately, current quality control methods for crowdsourcing consider only the question difficulty to account for the variation. As a result, these methods cannot distinguish workers' \textit{personal preferences} for different correct answers of a \textit{partially subjective} question 
from their \textit{ability/expertise} to avoid objectively wrong answers for that question. To address this issue, we present a probabilistic model which (i) explicitly encodes question difficulty as a model parameter and (ii) implicitly encodes question subjectivity via latent \textit{preference factors} for crowd-workers. 
We show that question subjectivity induces grouping of crowd-workers, revealed through clustering of their latent preferences. Moreover, we develop a quantitative measure of the subjectivity of a question. 
Experiments show that our model (1) improves the performance of both quality control for crowd-sourced answers and next answer prediction for crowd-workers, 
and (2) can potentially provide coherent rankings of questions in terms of their difficulty and subjectivity, so that task providers can refine their designs of the crowdsourcing tasks, e.g. by removing highly subjective questions or inappropriately difficult questions. 
\end{abstract}



%
%
\maketitle

\section{Introduction}
Outsourcing tasks to a flexible online workforce (aka crowdsourcing) has proven a successful paradigm for data collection in numerous fields due primarily to its overall lower costs and shorter turnaround time as compared to in-house expert-based data collection. The downside of online crowdsourcing is that the quality of the answers collected from crowd-workers is usually not guaranteed, even when multiple responses are collected and aggregated for each question, and workers are trained and vetted using gold-standard questions. To address this issue, many quality control methods for the crowdsourced answers have been proposed \cite{whitehill:nips09,welinder2010multidimensional}. These methods rely on the assumptions that most crowd-workers are reliable when answering the questions and that a given worker is more likely to be reliable should she agree with the majority of her co-workers on the majority of their jointly answered questions. Thus, the methods have focused on modelling the \emph{accuracy/ability/expertise} of individual workers, assuming this to be correlated with the quality of the responses \cite{dawid1979maximum,raykar2010learning}. In recent years, it has become popular for quality control methods to also model the influence that individual questions exert on the quality of the responses \cite{whitehill2009whose,bachrach2012grade}. Broadly speaking, the following two key properties of questions have drawn the modelling attention:
\begin{itemize}[leftmargin=10pt,topsep=0pt]
\item \textbf{Difficulty.} The modelling of question difficulty is founded on the assumption that greater agreement on workers' answers to a particular question indicates less difficulty for them in determining the correct response. Quality control methods often encode this assumption using a function in which worker expertise counteracts question difficulty for predicting the probability of a correct response. The probability is also known as the quality of the response: the more difficult the question, the lower the quality of a response, and vice versa. In addition, some methods (e.g. \cite{kamar2015identifying}) also consider the existence of \emph{deceptive} questions which are so difficult that the assumption that the majority of worker responses are correct no longer holds.
\item \textbf{Subjectivity.} In crowdsourcing, there are also many tasks that contain (\textit{purely} or \textit{partially}) subjective questions \cite{nguyen2016probabilistic}. Intuitively, the degree of subjectivity of a question (or equivalently, the data item described by it) depends on the number of answer options that are correct. Being purely subjective means all of the options are correct, while being partially subjective means more than one but not all of them are correct. Unless it is explicitly announced by the task provider that a question accepts all options (e.g. movie rating by workers to build a movie recommender system \cite{lee2013alleviating}), the number of correct answers to a question is unknown and assumed by most quality control methods to be one, meaning the question is objective. 
However, it is widely known that even expert assessors can disagree with each other on the correct answer to a question in typical crowdsourcing tasks like relevance judgement which is deemed to be ``quite subjective" (or equivalently, at least partially subjective) \cite{voorhees2000variations}. In this case, the objectivity assumption on the questions does not hold and most of the quality control methods based on this assumption cannot distinguish the answering \textit{accuracy/quality} of workers from their \textit{preferences} for the different answers of questions. 
 
\end{itemize}

\begin{figure*}
\centering
\begin{subfigure}[b]{0.9\columnwidth}
\centering
\includegraphics[width=1.6in]{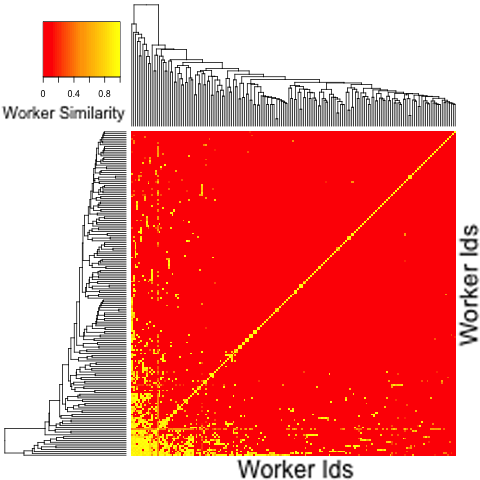}
\caption{\label{fig:worker_sim_obj} \emph{Product Matching task:} crowd-workers asked whether two product descriptions referred to the same item or not.}
\end{subfigure}%
\hspace{1.5cm}
\begin{subfigure}[b]{0.9\columnwidth}
\centering
\includegraphics[width=1.6in]{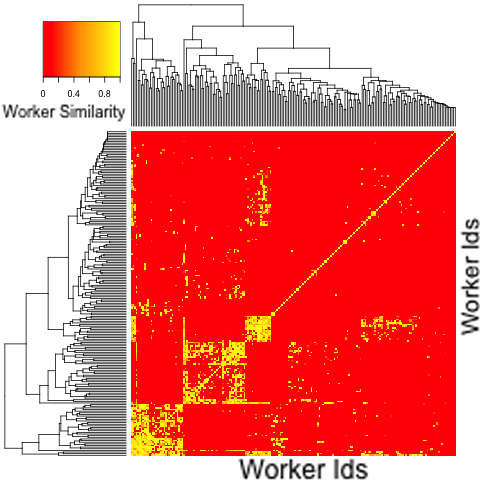}
\caption{\label{fig:worker_sim_sbj} \emph{Fashion Judgement task:} crowd-workers asked whether a picture contains a ``fashion related item'' or not.}
\end{subfigure}%
\caption{Heatmaps showing inter-worker response similarity (\% of response agreement) for two different tasks: (a) a \emph{relatively objective} product matching task and (b) a \emph{more subjective} fashion judging task, both involving binary worker responses. Hierarchical clustering was performed to order workers such that similar workers are close together. The three yellow blocks in the figure for task (b) indicate three groups of response behaviour and higher subjectivity for task (b).}
\label{fig:worker_sim}
\end{figure*}

For crowdsourcing tasks that contain questions whose subjectivity is either unknown or known to be at least partial, novel quality control methods need to be developed to capture any underlying answering/labelling pattern that results from the subjectivity of the questions. One such pattern uncovered by collaborative filtering \cite{koren2009matrix} is that crowd-workers who share similar preferences tend to respond similarly towards subjective questions which share certain (latent) features.
This pattern can also be observed in crowdsourcing where groups emerge amongst crowd-workers in terms of the answers they give to partially subjective questions. Figure \ref{fig:worker_sim} illustrates this phenomenon by providing heat maps of pairwise worker similarity for two tasks: (a) a relatively objective task and (b) a more subjective one. The objective task required workers to judge whether a pair of products were the same based on their names, descriptions and prices, while the subjective task asked workers to judge whether an image contained ``fashion related items''\footnote{The datasets for the two tasks have been listed in section \ref{sec:experiment}.}. The similarity between pairs of workers is calculated as the percentage of agreement across the jointly answered questions\footnote{Pairs of crowd-workers not sharing any items had their similarity to be $.00001$. 
} 
and hierarchical clustering has been performed to group similar workers together. The three yellow boxes along the diagonal for the more subjective task (b) indicates the three distinct groups of worker response behaviour for this task, which was absent in the more objective task (a).
Since the workers were mostly reliable on both tasks
, we conjecture that the grouping of response behaviour for the workers in the fashion judgement task reflects the underlying structures in their tendencies for selecting the different correct answers of the same questions (due to their subjectivity). 

To enable the answer quality control for the above tasks and generally, any crowdsourcing task that exhibits arbitrary degrees of question subjectivity and difficulty, we are motivated to develop a statistical model encoding both these properties. The resulting model is able to explain both the randomness and the correlations in the answering behaviour of crowd-workers. More specifically, when a task contains only \textit{purely subjective} questions, groupings of workers start to emerge due to the subjectivity of questions. A group captures a particular correlation between the crowd-workers within it and the latent correct answers for each of the questions. We model such a correlation by factorising it into the latent preferences of the workers and the latent features of the questions. The assumption is that the workers with similar latent preferences tend to have similar perceptions of what constitutes a correct answer for each of the questions. For instance, asking workers ``which colour for this shirt do you like?'' is a purely subjective question for which one group of workers who like blue colour in general will answer ``blue'', whereas another group who like green colour will choose ``green''. There is no reason to believe one of these two groups answer the question more correctly than the other, and their distinct answering patterns tend to remain consistent across similar questions asking about their colour favourites for other items (e.g. trousers, hats).


If a question is \emph{partially subjective}, this means it possesses (\textit{i}) a certain degree of subjectivity which corresponds to either its \textit{tendency} of having two or more correct answers, and (\textit{ii}) a certain level of difficulty. This difficulty corrupts the crowd-workers' perceptions as to (what tend to be) the correct answers of the question determined by its subjectivity to various extents depending on its level against the workers' levels of expertise. We model a greater extent of the corruption as a lower probability that the worker's answer is equal to the subjective (worker-specific) correct answer to the question, thereby the lower quality of the worker's answer. This subjective correct answer characterizes the particular \textit{group} to which the worker belongs by sharing similar \textit{preferences} with some other workers\footnote{We refer the reader to the movie example in the previous paragraph.}.

In this paper we introduce a new quality-control framework for crowdsourcing that \emph{models both the subjective (i.e. worker-specific) truths regarding the correct answers to individual questions and also the difficulty-dependent probability that a worker's answer to a question will equal her perceived subjective truth}. 
We now summarise the contributions of the paper as follows: 
\begin{itemize}[leftmargin=*,topsep=0pt]
    \item A novel statistical model is proposed which encodes the question difficulty explicitly and the question subjectivity implicitly via latent variables for worker preferences and corresponding question features. The model accounts for both the random and the systematic parts of the variance in crowdsourced answers to refine the quality control over them.
    \item A Monte Carlo simulation approach is provided for quantifying question subjectivity as the expected number of subjective truths perceived by different groupings of crowd-workers with respect to their preferences. 
    \item A meaningful ranking of questions in terms of either difficulty or subjectivity is derived from the model parameter estimates. This can bring practical benefits to crowdsourcing such as improving designs of tasks by helping requesters to detect and remove highly subjective questions from the tasks intended to be objective.
\end{itemize}

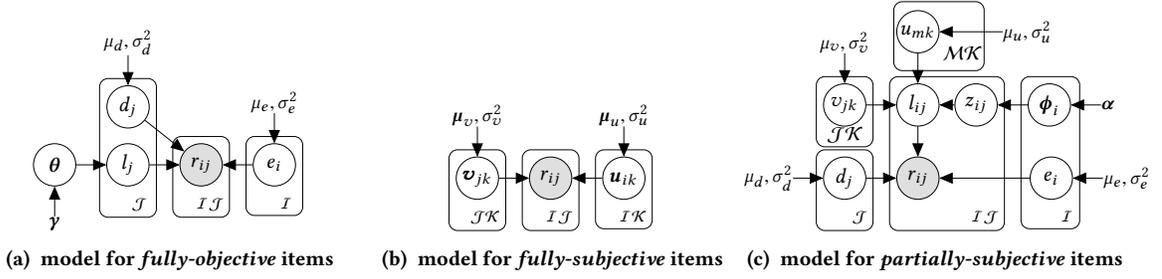
\begin{figure*}
  \begin{subfigure}[b]{0.6\columnwidth}
  \centering
  	\scalebox{.8} 		{\begin{tikzpicture}
  \node[obs] (rij) {$r_{ij}$};
  \node[latent, left=of rij, xshift=.5cm] (lj) {$l_{j}$};
  \node[latent, left=of lj, xshift=.5cm] (rho) {$\boldsymbol{\theta}$};
  \node[const, below=of rho, yshift=.5cm] (lambda) {$\boldsymbol{\gamma}$};
  \node[latent, right=of rij, xshift=-.5cm] (ei) {$e_{i}$};
  \node[latent, above left=of rij, yshift=-.25cm] (dj) {$d_{j}$};
  \node[const, above=of ei, yshift=-.5cm] (mu_sigma_ei) {$\mu_e,\sigma^2_e$};
  \node[const, above=of dj, yshift=-.5cm] (mu_sigma_dj) {$\mu_d,\sigma^2_d$};

    \edge{ei,dj,lj}{rij};
    \edge{mu_sigma_dj}{dj};
    \edge{mu_sigma_ei}{ei};
  \edge {rho} {lj};
  \edge {lambda} {rho};
  

  \plate {Response} {(rij)} {$\mathcal{IJ}$};
  \plate {Worker_Class} {(ei)}{$\mathcal{I}~~~~~~~~~~$}
  \plate {Item_Class} {(lj)(dj)}{$\mathcal{J}~~~~~~~~~~$}

\end{tikzpicture}}
    \caption{\label{fig:latent_var_model_true_answers} model for \emph{fully-objective} items}
  \end{subfigure}%
  \begin{subfigure}[b]{0.6\columnwidth}
	\centering
  	\scalebox{.8} 		{\begin{tikzpicture}
  \node[obs] (rij) {$r_{ij}$};
  \node[latent, right=of rij, xshift=-.5cm] (ui) {$\boldsymbol{u}_{ik}$};
  \node[latent, left=of rij, xshift=.5cm] (vj) {$\boldsymbol{v}_{jk}$};
  \node[const, above=of ui, yshift=-.5cm] (mu_sigma_ui) {$\boldsymbol{\mu}_u,\sigma^2_u$};
  \node[const, above=of vj, yshift=-.5cm] (mu_sigma_vj) {$\boldsymbol{\mu}_v,\sigma^2_v$};

    \edge{ui,vj}{rij};
    \edge{mu_sigma_vj}{vj};
    \edge{mu_sigma_ui}{ui};
  

  \plate {Response} {(rij)} {$\mathcal{IJ}$};
  \plate {Worker_Class} {(ui)}{$\mathcal{IK}$}
  \plate {Item_Class} {(vj)}{$\mathcal{JK}$}

\end{tikzpicture}}
    \caption{\label{fig:latent_var_model_no_true_answers} model for \emph{fully-subjective} items}
  \end{subfigure}%
  \begin{subfigure}[b]{0.6\columnwidth}
  	\centering
  	\scalebox{.8} 		
    {\begin{tikzpicture}


  \node[obs] (rij) {$r_{ij}$};
  \node[latent, above=of rij, yshift=-.5cm] (lij) {$l_{ij}$};
  \node[latent, right=of lij, xshift=-.75cm] (zij) {$z_{ij}$};
  \node[latent, right=of zij, xshift=-.5cm] (thetai) {$\boldsymbol{\phi}_{i}$};
  \node[const, right=of thetai, xshift=-.5cm] (alpha) {$\boldsymbol{\alpha}$};
  \node[latent, right=of rij, xshift=.5cm] (ei) {$e_{i}$};
  \node[latent, left=of rij, xshift=.5cm] (dj) {$d_{j}$};
  \node[latent, above=of lij, yshift=-.5cm] (um) {$u_{mk}$};
  \node[latent, left=of lij, xshift=.5cm] (vj) {$v_{jk}$};
  \node[const, right=of um, xshift=.05cm] (mu_sigma_um) {$\mu_u,\sigma^2_u$};
  \node[const, above=of vj, yshift=-.5cm] (mu_sigma_vj) {$\mu_v,\sigma^2_v$};
  \node[const, left=of dj, xshift=.5cm] (mu_sigma_dj) {$\mu_d,\sigma^2_d$};
  \node[const, right=of ei, xshift=-.5cm] (mu_sigma_ei) {$\mu_e,\sigma^2_e$};
  \node[const, right=of um, xshift=-.95cm, yshift=-.35cm] (mk) {$\mathcal{MK}$};
  \node[const, below=of vj, yshift=1cm] (vk) {$\mathcal{JK}$};
  \edge{alpha}{thetai}
  \edge{thetai}{zij}
	\edge{um, vj, zij}{lij};
    \edge{ei,dj,lij}{rij};
    \edge{mu_sigma_vj}{vj};
    \edge{mu_sigma_um}{um};
    \edge{mu_sigma_dj}{dj};
    \edge{mu_sigma_ei}{ei};
  

  \plate {Response} {(zij)(lij)(rij)} {$\mathcal{IJ}$}
  \plate [yshift=0.1cm, inner sep=1pt] {Worker_Class} {(um)(mk)}{}
\plate {Worker} {(ei)(thetai)}
{$\mathcal{I}$}
  \plate {Item_Class} {(vj)}{}
    \plate {Item} {(dj)}{$\mathcal{J}$}

\end{tikzpicture}}
    \caption{\label{fig:proposed_model1} model for \emph{partially-subjective} items}
  \end{subfigure}%
  \caption{(a) shows GLAD with a latent variable $l_j$ for each objective truth, (b) shows a collaborative filtering model without objective truths, and (c) is the proposed \textit{subjectivity-and-difficulty response} (SDR) model for partially-subjective questions that is able to distinguish question difficulty from subjectivity.}
\end{figure*}

\section{Related Work}
\subsection{Latent variable modelling in crowdsourcing}\label{section_latent_class_crowd}
Most state-of-the-art answer/label quality control methods in crowdsourcing have operated under the assumption that each question is purely objective. These methods are primarily based on statistical modelling of the interactions between crowd-workers and questions which determine either the marginal probabilities of the workers' answers equal to the corresponding correct answers \cite{whitehill2009whose,bachrach2012grade,welinder2010multidimensional} or the conditional probabilities of the answers given the correct answers \cite{dawid1979maximum,venanzi2014community,kamar2015identifying}. 
In comparison, the marginal probabilistic modelling is simpler than the conditional modelling, and also better at mitigating answer sparsity problem in crowdsourcing \cite{jung2013crowdsourced,jung2014quality}. The basic marginal probabilistic modelling is GLAD \cite{whitehill2009whose}, which models the correctness of each answer as a logistic function where the question difficulty counteracts the expertise of the responding worker. Its graphical representation is shown by Figure \ref{fig:latent_var_model_true_answers} with the following generative scheme for a response $r_{ij}$ of worker $i$ given to question $j$: $\boldsymbol{\theta} \sim Dir(\boldsymbol{\gamma});\;l_j \sim Discrete(\boldsymbol{\theta})\label{sample:true_answer};\text{   }\;r_{ij} \sim Discrete(\boldsymbol{\pi}_{ij})$. This means a correct answer $l_j$ is drawn for question $j$ from a discrete distribution parametrised by $\boldsymbol{\theta}$, which was previously drawn from a Dirichlet distribution parametrised by $\boldsymbol{\gamma}$. Then, a response $r_{ij}$ conditioned on $l_j$ is drawn from a discrete distribution with the $k$-th component of its parameters $\boldsymbol{\pi}_{ij}$ calculated as follows:
{\scriptsize
\begin{eqnarray}
\;\;\pi_{ijk} = f(e_{i}, d_{j})\;\; \text{if}\;\;k=l_j\;\;\text{else}\;\;\frac{1-f(e_{i}, d_{j})}{K-1};\;\;f(e_i, d_j) = \frac{1}{1+e^{-(e_i/\exp(d_j))}}
\label{equation:response_prob}
\end{eqnarray}
}In this case, the function takes in the expertise factor $e_i$ of worker $i$ and the difficulty factor $d_j$ of question $j$. The output of the function is the probability of the response $r_{ij}$ being correct. When $e_i \rightarrow +\infty$ or $d_j \rightarrow 0$, this probability grows, indicating a stronger positive correlation between $r_{ij}$ and $l_j$. When $e_i \rightarrow 0$ or $d_j \rightarrow +\infty$ and the question has binary options, the probability approaches 0.5, leading to no correlation between the two, which suggests $r_{ij}$ is a random binary pick. When $e_i \rightarrow -\infty$, the probability decreases to 0, indicating a stronger negative correlation. 

Although efficient in quality control of answers to objective questions, current models based on marginal probabilistic modelling have hardly considered modelling the subjectivity of questions, let alone inferring their possible subjective truths. One of the only two papers that have made progress in this regard is \cite{tian2012learning}. It assumes that a higher (lower) joint degree of difficulty and subjectivity for an entire crowdsourcing task can increase (decrease) the number of groups of answers given by the crowd-workers to the questions. The expected size of each group becoming smaller (larger) indicates overall weaker (stronger) correlations of answers given to the questions. Despite attributing the variance of answers to both difficulty and subjectivity, the paper makes no attempt to separate the two when it is supposed to be only the difficulty accounting for the quality of answers. Moreover, this work requires every question in a task to be answered by every worker, which is unrealistic in practice. The other work \cite{nguyen2016probabilistic} has focused on modelling partially subjective questions with just ordinal answers. It assumes each response to a question is generated by a Normal distribution the mean and the variance of which are linearly regressed over the observed features of the question. This means the model will poorly fit any multi-modal distribution of answers to a question.

\subsection{Latent variable modelling in collaborative filtering}
\label{section_latent_class_collaborate}

In model-based collaborative-filtering \cite{koren2009matrix}, matrix factorization is typically applied to predicting ordinal ratings provided by users to items (e.g. movies, songs). Its categorical version, shown in Figure \ref{fig:latent_var_model_no_true_answers}, is less commonly applied but is important for the construction of our model for the quality control of crowdsourced categorical answers. It has a generative process for the response $r_{ij} \sim Discrete(\boldsymbol{\psi}_{ij}),\;\text{where}\;\boldsymbol{\psi}_{ij} = \{\psi_{ijk}\}_{k \in \mathcal{K}}$ and $\mathcal{K}$ is the set of answer options, with its $k$-th component calcuated as:
{\scriptsize
\begin{eqnarray}
\psi_{ijk} = P(r_{ij} = k|\boldsymbol{U}_i, \boldsymbol{V}_j) = \exp(\boldsymbol{u}^{T}_{ik}\boldsymbol{v}_{jk}) / \sum_{k'\in\mathcal{K}}\exp(\boldsymbol{u}^{T}_{ik'}\boldsymbol{v}_{jk'})
\label{equation:soft_max}
\end{eqnarray}
}Here, $\boldsymbol{\psi}_{ij}$ is also called the \textit{soft-max} function, $\boldsymbol{u}_{ik}$ and $\boldsymbol{v}_{jk}$ are respectively the latent preferences of worker $i$ and the latent features of item $j$ in relation to the $k$-th answer option. The inner product term $\boldsymbol{u}^{T}_{ik}\boldsymbol{v}_{jk}$ indicates how much tendency worker $i$ responds to item $j$ with the $k$-th answer option.

\section{Proposed model}\label{section_our_model}
Our proposed model endeavours to combine the key characteristics of the latent variable models specified in section \ref{section_latent_class_crowd} and section \ref{section_latent_class_collaborate}. We call it \textit{SDR} model (Subjectivity-and-Difficulty Response model), which comprises an \emph{upstream module} which generates a \textit{subjective} truth for a question based on the worker's perception of the correct answer, and a \emph{downstream module} which imposes a \textit{difficulty}-dependent corruption on the subjective truth for generating the actual response from the worker to the question.
More specifically, in the upstream module, the latent subjective truth $l_{ij}$ of question $j$ as perceived by crowd-worker $i$ is drawn from a soft-max function specified by Eq. (\ref{equation:soft_max}) except that the original $r_{ij}$ in the equation is now replaced by $l_{ij}$. This function explains how the worker's latent preferences interact with the question's latent features to generate the subjective truth behind her response to the question. In the downstream, conditioned on the latent subjective truth $l_{ij}$, the response $r_{ij}$ actually given by worker $i$ to question $j$ is determined by the logistic function $f(e_i, d_j)$. It encodes how the worker expertise $e_i$ counteracts the question difficulty $d_j$ to corrupt the subjective truth into the response, which will be defined later in this section. Essentially, the above perception-corruption process is a generalisation of the corruption process of the correct answer signals from objective questions modelled in \cite{welinder2010multidimensional} by additionally considering the question subjectivity.


Unfortunately the upstream+downstream model described above suffers from an over-parameterisation issue whereby \emph{both} the upstream component (which determines the worker-specific correct answer) and the downstream component (which determines the noise resulting from worker inaccuracy) can \emph{independently and adequately} explain the variance observed in worker responses to the same question. In other words, the varied responses from different workers to the same question could equally be due to different perceptions on what constitutes the correct answer to the question or to difficulty of the question causing low accuracy amongst the respondents. To remedy this situation we explicitly enforce a group structure over workers in order to limit the variation in the perceptions across workers. This is done by changing the upstream module to have \textit{sparsity-inducing priors} over the latent preferences of crowd-workers. In this paper, we use the \textit{Latent Dirichlet Allocation} (LDA) \cite{blei2003latent} as such priors. The final graphical representation of the SDR model is shown in Figure \ref{fig:proposed_model1}.
The new upstream module of our model assigns a probability vector $\boldsymbol{\phi}_i$, which follows a Dirichlet with a concentration parameter $\boldsymbol{\alpha}$, to each worker $i$. Each component $\phi_{mi}$ of this probability vector reflects the worker's tendency of showing a particular preference $m$ among the set of preferences $\mathcal{M}$ she possesses when answering any question. Then, a preference assignment $z_{ij}$ is drawn from $\boldsymbol{\phi}_i$ for determining the specific preference worker $i$ will show for answering question $j$. As for preference $m$, it has a weight $u_{mk}$ for each answer option $k$ to reflect how likely each option is to be selected given the preference $m$ showed by any worker. In this paper, we fix the dimension of $u_{mk}$ to be strictly 1. This weight is multiplied with the latent feature $v_{jk}$ of question $j$ and the result is input to a soft-max function for drawing the subjective truth $l_{ij}$ behind the response $r_{ij}$. The above generative process can be formulated as: $\boldsymbol{\phi}_{i} \sim Dir(\boldsymbol{\alpha});\;\;z_{ij} \sim Discrete(\boldsymbol{\phi}_i);\;\;l_{ij} \sim Discrete(\boldsymbol{\psi}_{z_{ij}})$ with the $k$-th component of the soft-max function $\boldsymbol{\psi}_{z_{ij}}$ calculated as:{\scriptsize
\begin{eqnarray}
\psi_{z_{ij}k} = \frac{\exp(u_{z_{ij}k}v_{jk})}{ \sum\limits^{K}_{k'=1}\exp(u_{z_{ij}k'}v_{jk'})}
\label{sample:response_our_model}
\end{eqnarray}
}Embodying the sparsity-inducing effect of LDA, the preference probabilities $\boldsymbol{\phi}_i$ are dedicated to revealing the underlying groups of crowd-workers while the soft-max specified by Eq. (\ref{sample:response_our_model}) governs the positive correlations between the latent correct answers to the same questions perceived by the workers within the same group. When the number of preferences in $\mathcal{M}$ = 1, the probability of the only preference $\phi_i$ is 1. This has a two-fold meaning that each question has one correct answer and every worker should perceive the correct answer of any question in the same way. When the size of $\mathcal{M}$ is greater than 1, this indicates certain numbers of underlying worker groups, which we can recover by applying K-means clustering to the estimated preference probabilities $\hat{\boldsymbol{\phi}}_i$ using the Elbow method to determine the right number of the groups.

The downstream module corrupts the correlations between the subjective truth $l_{ij}$ and the response $r_{ij}$. It draws $r_{ij}$ from a discrete probability distribution $\boldsymbol{\pi}_{ij}$ specified in Eq. (\ref{equation:response_prob}) except the logistic function $f(e_i, d_j)$ has the following definition from \cite{rasch1993probabilistic}:
{\scriptsize
\begin{eqnarray}
f(e_i, d_j) = \frac{1}{1+e^{-(e_i - d_j)}}
\label{equation:correctness_prob}
\end{eqnarray}
}The term $(e_i - d_j)$ naturally explains the type of biases induced by \textit{deceptive} questions when the difficulty $d_j$ is much larger than the expertise $e_i$, which is not captured in Eq. (\ref{equation:response_prob}) as the term $\exp(d_j)$ is never smaller than 0, meaning questions never bias workers to answer incorrectly due to their difficulty. Moreover, when the estimated values for this term are greater than zero for most responses, it means SDR deems them more likely to be correct. With more of them deemed correct, the number of inferred correct answers to any question tends to increase. As a result, the size of latent preference set $\mathcal{M}$ should grow, from the perspective of SDR, to fit the seemingly more diverse set of correlations between latent correct answers across the questions. Thus, for our model to recover the right number of latent preferences for crowd-workers from their responses, the priors for $e_i$ and $d_j$ need to be set properly, which will be elaborated more in section \ref{section:hyperparam_setup}. 

\section{Estimation}
\subsection{Model parameter estimation}
We now provide equations used for parameter estimation, using the notation $\psi_{z_{ij}k}$ from Eq. (\ref{sample:response_our_model}) and $f(e_i, d_j) = f_{ij}$ from Eq. (\ref{equation:correctness_prob}) to simplify the equations. The conditional probability for the preference assignment $z_{ij}$ to worker $i$ when answering question $j$ given the model parameters is:
{\scriptsize
\begin{flalign}
P(z_{ij}\!=\!m|e_i,d_j,\boldsymbol{u}_{m},\boldsymbol{v}_{j},\alpha) \propto\sum\limits_{k\in\mathcal{K}}\psi_{mk}f^{\delta_{ijk}}_{ij}\bigg(\frac{1 - f_{ij}}{K-1}\bigg)^{1-\delta_{ijk}}  \frac{N^m_{i\neg{j}} + \alpha_m}{\sum\limits_{m\in\mathcal{M}}N^{m'}_{i\neg{j}} + \alpha_{m'}}
\label{sample:preference}
\end{flalign}}where $N^m_{i\neg{j}}$ denotes the number of questions excluding question $j$ answered by worker $i$ given her preference $m$. The joint probability of the other parameters given $z_{ij}$ and the hyper-parameters is:
{\scriptsize
\begin{multline}
Q = p\Big(\{e_i\}_{i\in\mathcal{I}},\{d_j,\boldsymbol{v}_j\}_{j\in\mathcal{J}},\{\boldsymbol{u}_{m}\}_{m\in\mathcal{M}}|\{\boldsymbol{z}_{ij}\}_{i\in\mathcal{I},j\in\mathcal{J}},\mu_{\{e,d,u,v\}},\sigma^2_{\{e,d,u,v\}}\Big) \\= - \sum\limits_{i\in\mathcal{I}}\sum\limits_{j\in\mathcal{J}}\log\Bigg[\sum\limits_{k\in\mathcal{K}}\Bigg(\psi_{z_{ij}k}f^{\delta_{ijk}}_{ij}\bigg(\frac{1 - f_{ij}}{K-1}\bigg)^{1-\delta_{ijk}}\Bigg)\Bigg] - \sum\limits_{i\in\mathcal{I}}\log\big(p(e_i|\mu_e,\sigma^2_e)\big) -\\ \sum\limits_{j\in\mathcal{J}}\log\big(p(d_j|\mu_d,\sigma^2_d)\big) - \sum\limits_{k\in\mathcal{K}}\sum\limits_{m\in\mathcal{M}}\log\big(p(u_{mk}|\mu_u,\sigma^2_u)\big)-\sum\limits_{j\in\mathcal{J}}\log\big(p(v_{jk}|\mu_v,\sigma^2_v)\big)
\label{equation:optimize}
\end{multline}}The partial derivatives for $Q$ with respect to the other parameters: 
{\scriptsize
\begin{flalign}
\frac{\partial Q}{\partial u_{mk}}=-\sum\limits_{i\in\mathcal{I}}\sum\limits_{j\in\mathcal{J}}\zeta_{ijm}v_{jk}\sum\limits_{k'\in\mathcal{K}}f^{\delta_{ijk'}}_{ij}\bigg(\frac{1 - f_{ij}}{K-1}\bigg)^{1-\delta_{ijk'}}\omega_{ijkk'} +\frac{u_{mk}-\mu_u}{\sigma^{2}_u}
\label{equation:partial_derivative_umk}
\end{flalign}
\begin{multline}
\frac{\partial Q}{\partial v_{jk}}=-\sum\limits_{i\in\mathcal{I}}u_{z_{ij}k}\sum_{k'\in\mathcal{K}}f^{\delta_{ijk'}}_{ij}\bigg(\frac{1 - f_{ij}}{K-1}\bigg)^{1-\delta_{ijk'}}\omega_{ijkk'}  +\frac{v_{jk}-\mu_v}{\sigma^{2}_v}
\label{equation:partial_derivative_vjk}
\end{multline}
\begin{multline}
\frac{\partial Q}{\partial e_{i}}=-\sum\limits_{j\in\mathcal{J}}\sum_{k\in\mathcal{K}}\bigg(\frac{-1}{K-1}\bigg)^{1-\delta_{ijk}}f_{ij}(1-f_{ij})\psi_{z_{ij}k} +\frac{e_{i}-\mu_e}{\sigma^{2}_e}  
\label{equation:partial_derivative_ei}
\end{multline}
\begin{multline}
\frac{\partial Q}{\partial d_{j}}=-\sum\limits_{i\in\mathcal{I}}\sum_{k\in\mathcal{K}}(-1)^{\delta_{ijk}}\bigg(\frac{1}{K-1}\bigg)^{1-\delta_{ijk}}f_{ij}(1-f_{ij})\psi_{z_{ij}k} +\frac{d_{j}-\mu_d}{\sigma^{2}_d}
\label{equation:partial_derivative_dj}
\end{multline}
}Here, $\delta_{ijk}=\mathbbm{1}{\{r_{ij} = k\}}, \zeta_{ijm}=\mathbbm{1}\{z_{ij} = m\}$ and $\omega_{ijkk'}=\psi_{z_{ij}k}\big(1-\psi_{z_{ij}k}\big)^{\mathbbm{1}\{k = k'\}}\big(-\psi_{z_{ij}k'}\big)^{\mathbbm{1}\{k \neq k'\}}$. The parameter estimation involves two alternating procedures: sample $z_{ij}$ according to Eq. (\ref{sample:preference}) and optimize $Q$ in Eq. (\ref{equation:optimize}) using LBFGS based on Eq. (\ref{equation:partial_derivative_umk}), (\ref{equation:partial_derivative_vjk}), (\ref{equation:partial_derivative_ei}) and (\ref{equation:partial_derivative_dj}).
\subsection{True answer estimation}\label{section:true_label_prediction}
A single worker-specific correct answer $l_{ij}$ (as perceived by worker $i$) fails to provide overall information about the correct answers to question $j$. Thus, we should gather the $l_{ij}$ values from all workers who answer each question. However, in practice, each question is assigned to only a limited number (usually 3 or 5) of workers, making the estimate of the true answer distribution poor. Our solution to improving this estimate is to first find underlying clusters of workers (across all questions) by applying K-means with the Elbow method based on 10-fold cross validation to the posterior means $\hat{\boldsymbol{\Phi}} = \{\hat{\boldsymbol{\phi}}_i | i \in \mathcal{I}\}$ of the latent preference probabilities of all the workers. With the centroid $\hat{\boldsymbol{\phi}}_{c}$ of each resulting cluster $c$, we then calculate the probability that the true answer $l_{cj}$ (as perceived by the workers in cluster $c$) takes the value $k$ as follows:
{\scriptsize
\begin{eqnarray}
\begin{split}
P(l_{cj} = k | c) = \sum_{m\in\mathcal{M}}P(l_{cj} = k | m)P(m | c) = \sum_{m\in\mathcal{M}}\Bigg(\frac{\exp(\hat{u}_{mk}\hat{v}_{jk})}{\sum_{k'=1}^{K}\exp(\hat{u}_{mk'}\hat{v}_{jk'})}\hat{\phi}_{mc}\Bigg)
\end{split}
\label{equation:group_true_label}
\end{eqnarray}
}where $\hat{u}_{mk}$ and $\hat{v}_{jk}$ are the estimates of the weight $u_{mk}$ for preference $m$ and the latent feature $v_{jk}$ of question $j$, both specific to option $k$. The best guess regarding the correct answer $l_{cj}$ according to the workers assigned to cluster $c$ is then:
{\scriptsize
\begin{eqnarray}
\hat{l}_{cj} = \arg\max_{k \in \mathcal{K}} P(l_{cj} = k | c)
\label{sample:cluster_true_answer}
\end{eqnarray}
}Now we have a set of correct answer estimates $\hat{\mathcal{L}}_j = \{\hat{l}_{cj} |c\in\mathcal{C}\}$ for question $j$ from all the worker clusters (with $\mathcal{C}$ being the set of the clusters). For the task of true answer prediction, we can either arbitrarily choose one from $\hat{\mathcal{L}}_j$ as the estimate of the correct answer $l_j$ or choose by following certain strategies. Two simple strategies are to choose $\hat{l}_{cj}$ from the cluster $c$ with the highest average expertise over its workers, or from the cluster with the largest proportion of workers assigned to it. The first strategy states that the correct answer perceived by on average the most expert group of workers is the most appropriate, while the second assumes it to be the one perceived by the largest group of workers which represents the mainstream school-of-thought. In this paper, we apply the second strategy because most crowdsourcing datasets used in the experiments correspond to relatively simple tasks, where the provided correct answers we believe are more likely to be mainstream opinions. As for the first strategy, it might be more useful than the second for revealing a minority group of expert workers who show distinct preferences on partially or purely subjective questions from the majority of less expert workers.

\subsection{Subjectivity estimation}
Despite not being directly estimated in the model, question subjectivity can still be quantified and estimated after the model has been estimated. This is achieved based on the reasonable assumption that the subjectivity of each question is proportional to the number of correct answers it affords. Despite not knowing the actual number of correct answers $|\mathcal{L}_j|$ to question $j$, we can estimate the value by taking its expectation with respect to the clusters of workers derived in section \ref{section:true_label_prediction}. More precisely, the expected number of correct answers to question $j$ with respect to worker clusters $\mathcal{C}$ is: $\mathbb{E}_{\mathcal{C}}[|\mathcal{L}_j|] = \sum_{n = 1}^{|\mathcal{K}|}nP_{\mathcal{C}}(|\mathcal{L}_j| = n)$. In this equation, $n$ iterates over the possible number of correct answers (from 1 to the size of $\mathcal{K}$). The probability $P_{\mathcal{C}}(|\mathcal{L}_j| = n)$ denotes how likely it is that the number of correct answers to question $j$ equals $n$, with respect to the worker clusters $\mathcal{C}$. It is, however, difficult to calculate this probability when $\mathcal{C}$ and $\mathcal{K}$ are large due to a combinatorial explosion. Thus we apply Monte Carlo simulation to estimate (a measure of) the subjectivity of question $j$ as $\mathbb{E}_{\mathcal{C}}[|\mathcal{L}_j|]$ using Alogrithm \ref{table:subj_est}.{\scriptsize
\begin{algorithm}[!htbp]
\caption{Subjectivity estimation for question $j$}
\label{table:subj_est}
\begin{algorithmic}[1]  
\Require $\hat{v}_j;\;\{\hat{\boldsymbol{u}}_{m}\}_{m\in\mathcal{M}};\;\; \hat{\boldsymbol{\Phi}}_c = \{\hat{\boldsymbol{\phi}}_c\}_{c\in\mathcal{C}};\;\;T = 50,000$.
\Ensure $\mathbb{E}_{\mathcal{C}}[|\mathcal{L}_j|]$.
\State $n_j \gets 0$.$\;$ {\color{blue} /* Initialise number of correct answers for question $j$ to zero */ }
\For{$t = 1 ... T$} {\color{blue} /* Sample over $T$ iterations.*/}
\State $\hat{\mathcal{L}_j} \gets \{\}$.$\;$ {\color{blue}/* Initialise set of correct answers to be sampled at iteration $t$. */ }
\For{$c = 1 ... C$}
\State $z_{cj} \sim Discrete(\hat{\boldsymbol{\phi}}_c)$.$\;$ {\color{blue} /* Sample group preference assignment $z_{cj}$. */}
\State $\hat{l}_{cj} \sim \boldsymbol{\psi}_{z_{cj}}, \;\text{where}\;\psi_{z_{cj}k} = \exp(\hat{u}_{z_{cj}k}\hat{v}_{jk}) / \sum\limits_{k'\in\mathcal{K}}\exp(\hat{u}_{z_{cj}k'}\hat{v}_{jk'})$.$\;$ {\color{blue}/* Sample correct answer $\hat{l}_{cj}$ perceived by worker cluster $c$. */}
\State $\hat{\mathcal{L}}_j \gets \hat{\mathcal{L}}_j \cup \hat{l}_{cj}\;$ only if $\;\hat{l}_{cj} \not\in \hat{\mathcal{L}}_j$ {\color{blue}/* Add sampled $\hat{l}_{cj}$ to $\hat{\mathcal{L}_j}$ when it first appears. */}
\EndFor
\State $n_{j} \gets n_{j} + |\hat{\mathcal{L}}_j|$. {\color{blue}/* Increase $n_j$ by number of distinct correct answers sampled at $t$. */}
\EndFor
\State $\mathbb{E}_{\mathcal{C}}[|\mathcal{L}_j|] \approx \frac{n_j}{T}$.$\;${\color{blue}/* Divide $n_j$ by $T$ to estimate $\mathbb{E}_{\mathcal{C}}[|\mathcal{L}_j|]$ as the question's subjectivity. */}
\end{algorithmic}
\end{algorithm}
}

\section{Experiments}\label{sec:experiment}
The evaluation of our proposed model consists of four parts. The first part is its sensitivity to various degrees of subjectivity in different crowdsourcing tasks. The second and the third parts are its performance of predicting respectively the provided correct answers of questions and the answers to be given by crowd-workers to unseen questions. The last part is its consistency with human assessors in assessing the difficulty and the subjectivity of questions. We have used 10 crowdsourcing datasets to evaluate the performance of our model in the experiments corresponding to the four parts. Table \ref{table:data_summary} summarises these datasets as being either (primarily) objective or partially subjective. Among them, the identification tasks of event time ordering, dog and duck breeds, and same products concern objective factual knowledge, while the judgement tasks of image beauty, document relevance 1\&2\footnote{The questions of relevance judgement task 2 come from the part of TREC 2011 crowdsourcing track \cite{lease2011overview} that does not contain the questions of relevance judgement task 1. We collected crowdsourced judgements for the task 2 from CrowdFlower.}, facial expression and adult content intrinsically contain certain degrees of subjectivity.

\subsection{SDR hyper-parameter setup}\label{section:hyperparam_setup}
As discussed at the end of section \ref{section_our_model}, to find the right number of latent preferences for crowd-workers, the hyper-parameters of the expertise $e_i$ and the difficulty $d_j$ in the SDR model need to be carefully set. This is achieved through \textit{held-out validation} which leverages \textit{noise} within worker responses for detecting signs that SDR may be overfitting the responses by introducing more latent preferences than necessary. More specifically, we construct a held-out validation dataset by randomly sampling a response from each worker. Thus, the size of such a dataset equals the number of workers participating in a task. Then, given a certain setting of the hyper-parameters, we learn our model based on the remaining responses and use the parameter estimates from the learned model to calculate the prediction accuracy: $1\!-\!MAE$ (Mean Absolute Error) over the held-out dataset. We repeat the model learning process with each hyper-parameter setting over the same 100 random held-out validation data subsets. We then obtain the average prediction accuracy for our model across these subsets for each hyper-parameter setting. Finally, we choose the setting (including the number for latent preferences) that yields the highest average prediction accuracy for use in the experiments.

\subsection{Sensitivity analysis}\label{section:sensitivity_analysis}
We first verify whether our model is sensitive to various degrees of subjectivity in different crowdsourcing tasks. If a task is (almost entirely) objective, the optimal size of latent preference set $\mathcal{M}$ should be 1, meaning that every crowd-worker now perceives the correct answers in the same way. Consequently, the probabilities of latent preferences $\boldsymbol{\phi}_i$ for worker $i$ collapse to $\phi_i = 1$, and the set of correct answers $\mathcal{L}_j$ for question $j$ collapses to a single correct answer $l_j$. In this case, we conduct the held-out validation on our model across the objective datasets each with the 100 randomly sampled data subsets described in section \ref{section:hyperparam_setup}. We expect that the average held-out prediction accuracy for our model across these data subsets will decrease when the number of latent preferences it has increases from 1 to 2, since in this case the model starts to overfit by learning the noise in the training responses to those objective tasks.

If a task is sufficiently subjective, our model should uncover the right number of underlying groups of workers along with the right number of latent preferences. We conduct the experiment in the same way as above to see the difference in average prediction accuracy on held-out unseen responses with the number of preferences increasing from 1 to 3 over the partially subjective datasets. We expect the average prediction accuracy to be higher when the number of preferences is greater than 1. Moreover, since Tian and Zhu~\cite{tian2012learning} provided us with the number of worker clusters emerging respectively from the five sub-tasks which constitute the image data in Table \ref{table:data_summary}, we thus compare the corresponding numbers of clusters derived from our model with theirs. 

\begin{table}[!t]
\centering
\scriptsize
\caption{The objective and the partially subjective datasets used in this paper.  \vspace{-.8cm}}
\begin{tabular}{|c|c|c|c|} 
\hline
\bf \textit{Objective datasets} &\bf \# Worker&\bf \# Item&\bf \# Response\\
\hline
\bf Time \cite{snow2008cheap} & 76 & 462 & 4,620\\ 
\bf Dog  \cite{NIPS2012_4490} & 109 & 807 & 8,070 \\
\bf Duck \cite{welinder2010multidimensional} & 53 & 240 & 9,600 \\
\bf Product \cite{wang2012crowder}  & 176  & 8,315 & 24,945 \\ 
\hline
\bf \textit{Partially subjective datasets} &\bf \# Worker&\bf \# Item&\bf \# Response\\
\hline
\bf Image \cite{tian2012learning} & 402 & 60 & 24,120\\
\bf Rel1  \cite{buckley2010overview} & 642 & 1,787 & 13,310\\ 
\bf Rel2 \cite{lease2011overview} & 83 & 585 & 1,755\\
\bf Fashion \cite{Loni:2013} & 199  & 3,837 & 11,511 \\ 
\bf Face \cite{DBLP:journals} & 27 & 584  & 5,242\\
\bf Adult \cite{adultjud}  & 269 & 333  & 3,324\\
\hline
\end{tabular}
\label{table:data_summary}
\end{table}

\subsection{Question correct answer prediction}\label{section:true_label_pred}
To verify the ability of the SDR model to predict the question true answers, we compare it with the following state-of-the-art quality control methods for crowdsourcing. All of these methods assume that each question has a single correct answer.
\begin{itemize}[leftmargin=*]
\item{\textbf{Majority Vote (MV)}}: The predicted correct answer for each question is the one chosen by the majority of the workers.
\item{\textbf{Multi-dimensional Wisdom of Crowds (MdWC) }\cite{welinder2010multidimensional}}: This model endows both crowd-workers and questions with multi-dimensional latent factors, and provides the workers with additional variables to account for their answering biases.
\item{\textbf{Generative model of Labels, Abilities, \& Difficulties} (\textbf{GLAD}) \cite{whitehill2009whose}:} This model resembles MdWC except that its latent factors (interpreted respectively as expertise and difficulty) are uni-dimensional, and it does not have worker-specific bias variables.
\item{\textbf{Dawid-Skene (DS)} \cite{dawid1979maximum}}: Unlike GLAD and MdWc which model the correctness probability of each worker's response, this model focuses on the (worker-specific) conditional probability of each response option given the correct answer to each question.
\item{\textbf{Community Dawid-Skene (CDS)} \cite{venanzi2014community}}: This model extends DS by clustering workers over some latent structures imposed on their conditional response probability matrices (given all correct answer possibilities) to alleviate the response sparsity problem.
\end{itemize}
The performance measure: \emph{correct answer prediction accuracy}, is calculated as:
$\frac{1}{|\mathcal{J}|}\sum_{j\in\mathcal{J}}\mathbbm{1}\{l_j=\hat{l}_j\}$, where $\hat{l}_j$ is inferred from the respective baselines. For our model, it is: $\frac{1}{|\mathcal{J}|}\sum_{j\in\mathcal{J}}\mathbbm{1}\{l_j=\hat{l}_{cj}\}$, where $c = \arg\max_{c\in\mathcal{C}} N_{c}$ with $N_{c}$ the number of workers assigned to cluster $c$ after Elbow K-means, and $\hat{l}_{cj}$ calculated by Eq. (\ref{sample:cluster_true_answer}). The hyper-parameters for each baseline except MV are optimised using the held-out validation specified in section \ref{section:hyperparam_setup} on the exact same random held-out validation subsets of each dataset in Table \ref{table:data_summary}.

\subsection{Worker answer prediction}\label{section:worker_ans_pred}
Predicting the answers to be given by crowd-workers to unseen questions is much more significant for (partially) subjective crowdsourcing tasks than it is for the objective ones as the former type of tasks values more about the different ways workers respond. For example, it is crucial to employ worker answer prediction to test a recommender system built on crowdsourced ratings. In this experiment, we evaluate the performance of all the models except MV on predicting the \textit{next answer} from each worker. We first sample one answer from each worker to create a \textit{held-out test} dataset, and then learn all the models from the rest of the data with their hyper-parameters optimised as described in section \ref{section:hyperparam_setup} using the exact same random validation data subsets. Finally, we evaluate the prediction performance of the models on the held-out test data using (1 - MAE). Due to the limitation of our computing power, in this experiment, we reduce the number of \textit{held-out validation} iterations for each model to be 15 before a single iteration of \textit{held-out test} is conducted. We perform 15 such random tests before the average performance of each model is elicited. 

\begin{figure}[t]
\begin{subfigure}[b]{0.236\textwidth}
\centering
\includegraphics[width=1.5in]{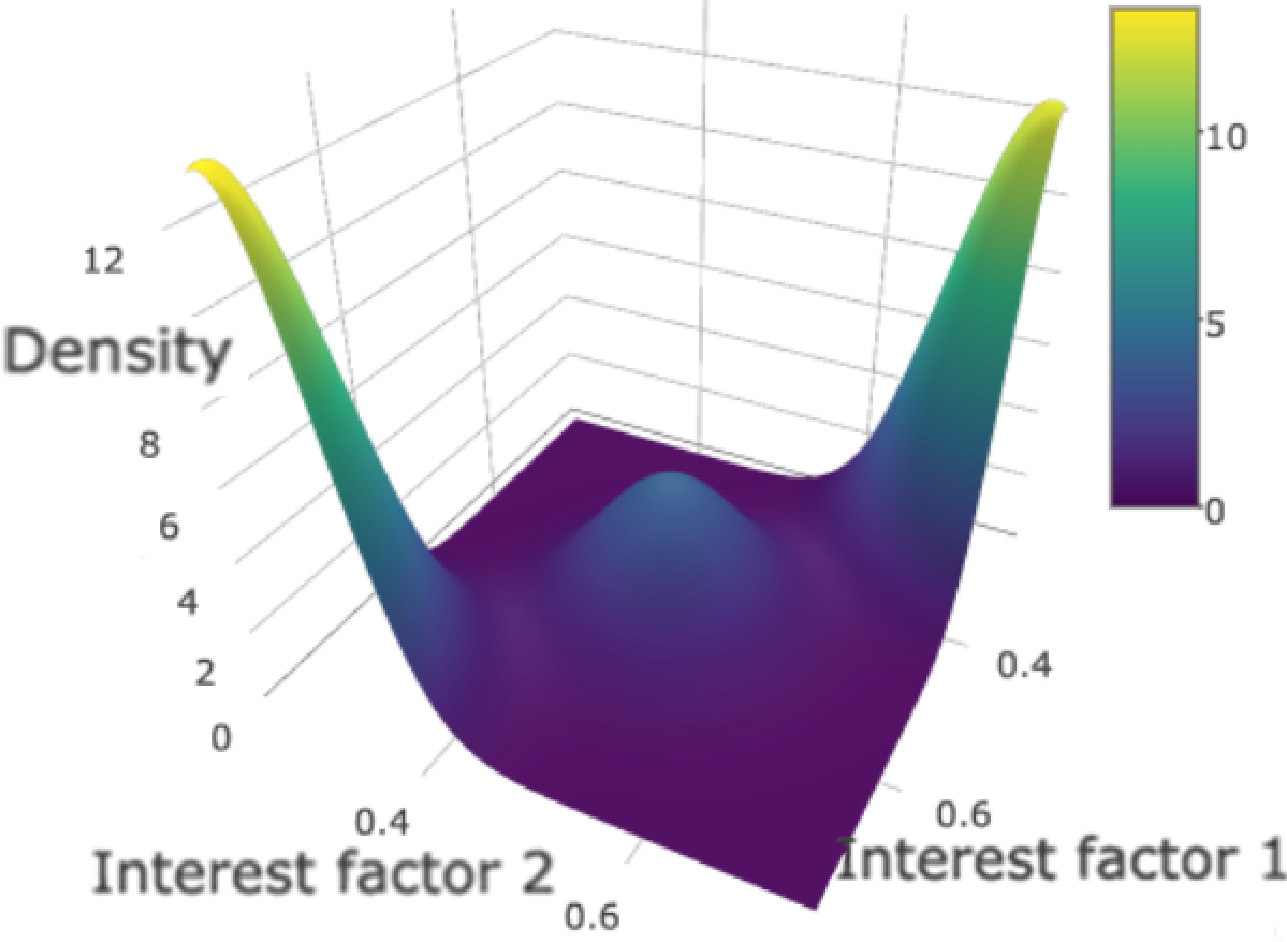}
\caption{}
\label{figure:density_plot_sky}
\end{subfigure} 
\begin{subfigure}[b]{0.236\textwidth} 
\centering
\includegraphics[width=1.5in]{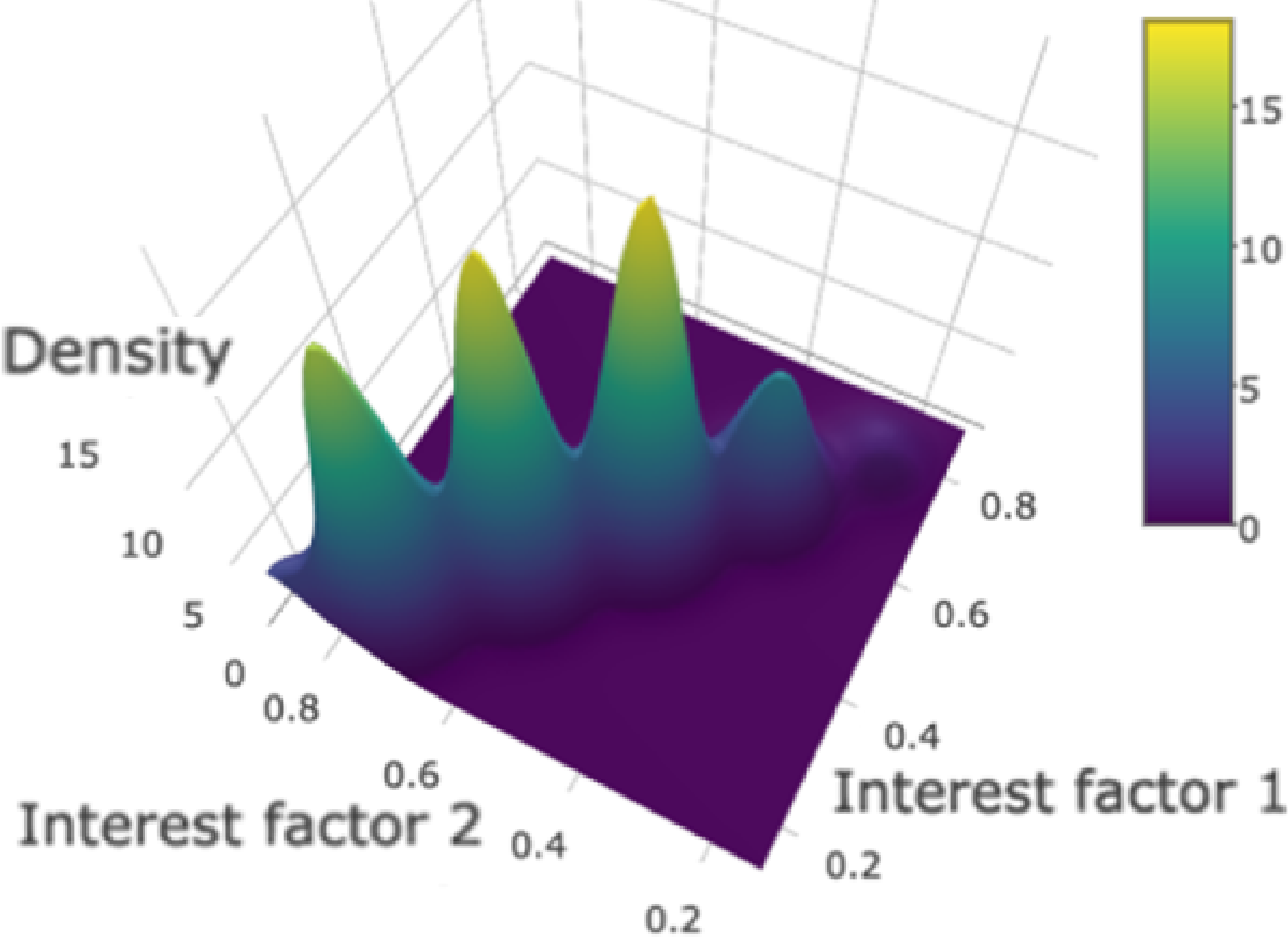}
\caption{}
\label{figure:density_plot_beauty}
\end{subfigure}%
\caption{(a) shows the 3 worker clusters on identifying sky from images and (b) shows the 4 worker clusters on judging beautiful images.}
\label{figure:density_plot}
\end{figure}

\subsection{Subjectivity and difficulty coherence}
In this experiment, we investigate whether the estimates of the difficulty and the subjectivity of questions derived from the SDR model are \textit{consistent} with the judgements of five human assessors. We focused on the object identification \& image aesthetics task\footnote{Crowd-workers are asked whether an image is beautiful or not.} from \cite{tian2012learning} as the total number of its questions is 60, a manageable workload for the assessors to provide good-quality judgements with sufficient levels of effort and concentration. The assessors are either PhD or Master students, three of whom are avid photographers with adequate knowledge about what constitutes beautiful images, while the other two are novices who, during the group discussion, provided suggestions as to how novices might react to different images. We ask them to rank the images with respect to (i) difficulty and (ii) subjectivity. The respective instructions we gave to them are:\textit{"rank all these images by how hard they are for crowd-workers to judge correctly by avoiding possible incorrect answers"} and \textit{"rank them this time by how subjective they are for crowd-workers to judge"}. The assessors first independently came up with their two rankings. In the process, they could redo the two ranking tasks until they felt confident to submit. The assessors then worked together to merge their rankings into single rankings (for both difficulty and subjectivity) through group discussion and majority vote. The resulting rankings were then compared with the corresponding rankings based on the estimates from the learned SDR model. In addition to ranking the images, the assessors were also asked to categorise each image into one of the three levels of difficulty (namely \emph{easy}, \emph{medium}, and \emph{hard}), and into one of the three levels of subjectivity (namely \emph{objective}, \emph{partially subjective}, and \emph{purely subjective}). We did this to see whether there existed any correlation between the difficulty or subjectivity levels to which images were categorised, and their corresponding estimates from the model.

\section{Results}

\begin{figure*}[t]
\centering
\begin{subfigure}[b]{0.3\textwidth}
\includegraphics[width=1.6in]{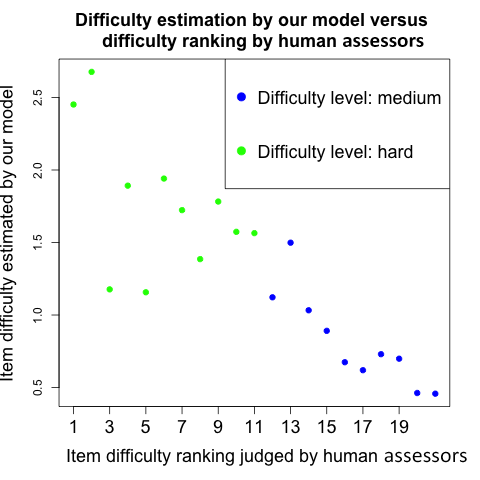}
\caption{}
\label{figure:a}
\end{subfigure}
\begin{subfigure}[b]{0.3\textwidth}
\includegraphics[width=1.6in]{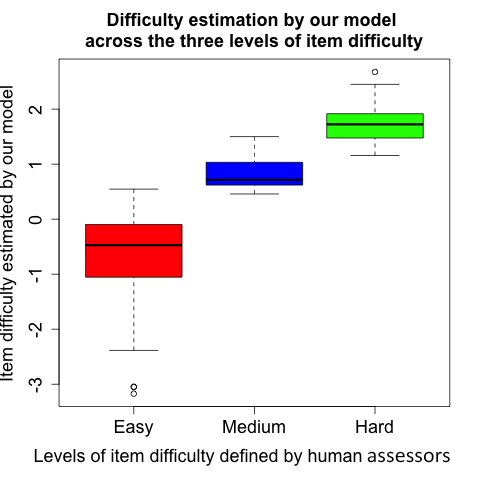}
\caption{}
\label{figure:b}
\end{subfigure}
\begin{subfigure}[b]{0.3\textwidth}
\includegraphics[width=1.6in]{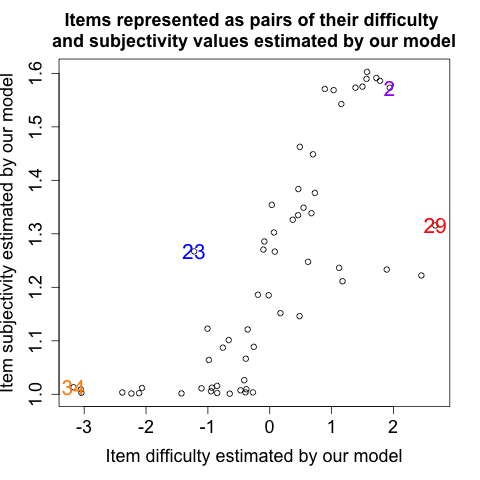}
\caption{}
\label{figure:c}
\end{subfigure}
\begin{subfigure}[b]{0.3\textwidth}
\includegraphics[width=1.6in]{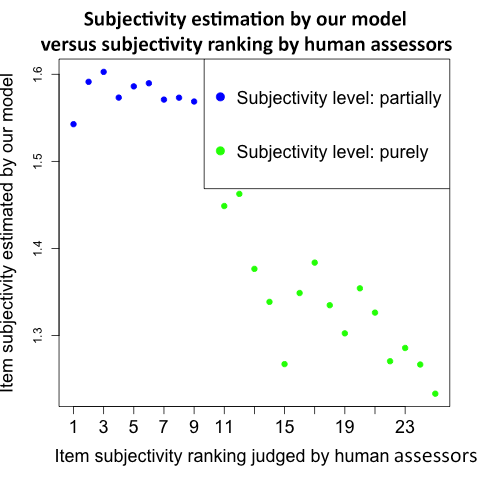}
\caption{}
\label{figure:d}
\end{subfigure}
\begin{subfigure}[b]{0.3\textwidth}
\includegraphics[width=1.6in]{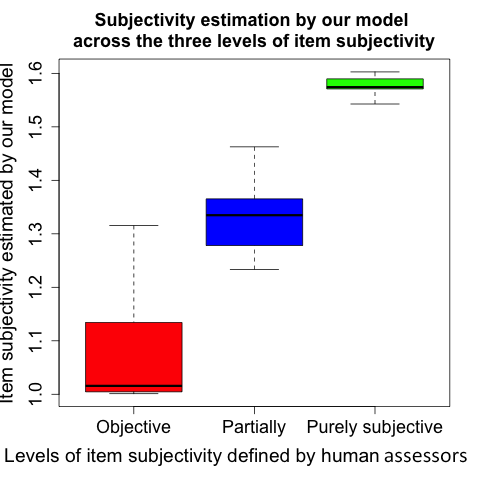}
\caption{}
\label{figure:e}
\end{subfigure}
\begin{subfigure}[b]{0.3\textwidth}
\includegraphics[width=1.6in]{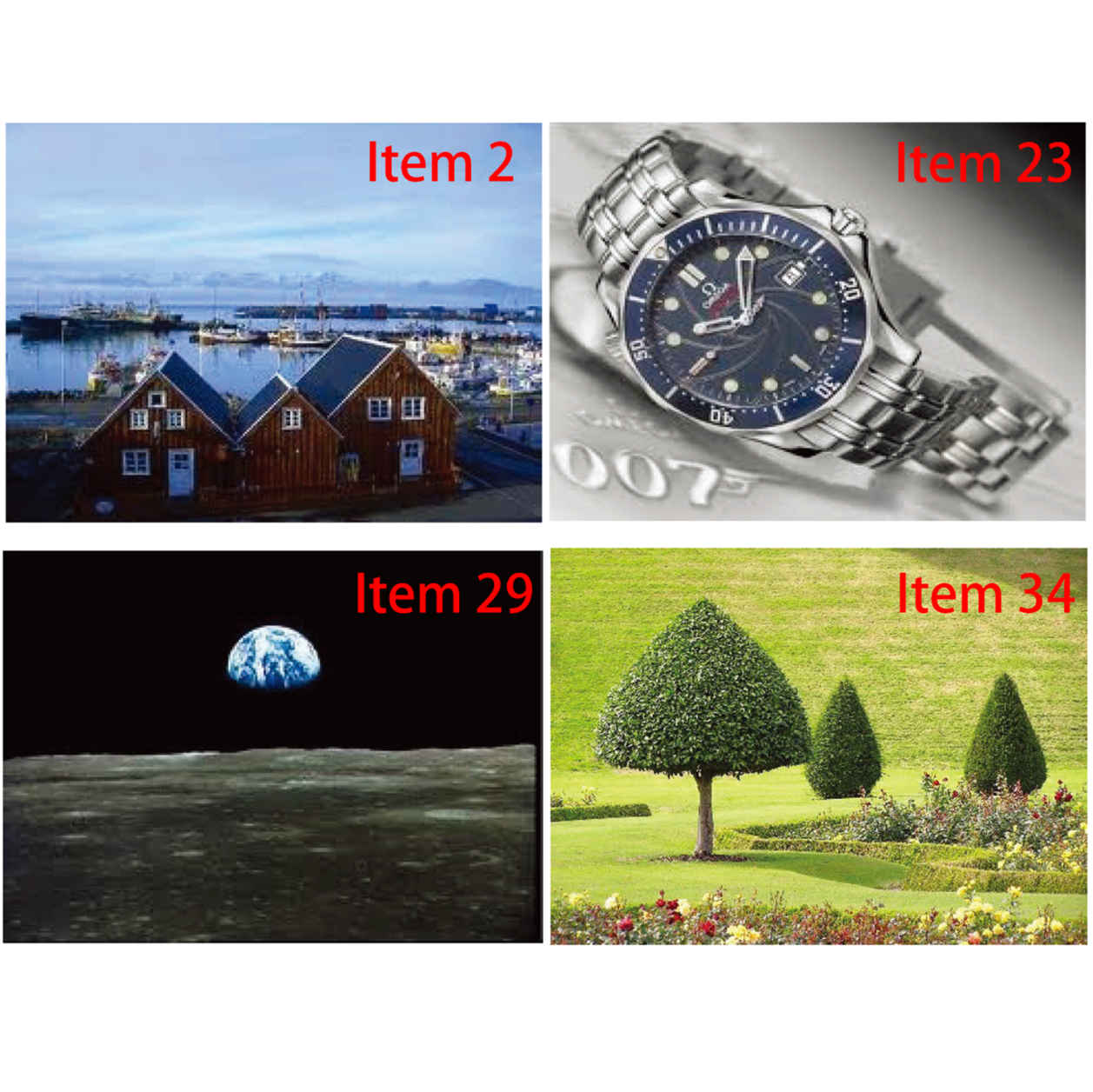}
\caption{}
\label{figure:f}
\end{subfigure}
\caption{ (a) and (d) show the correlation of the difficulty estimates and that of the subjectivity estimates respectively with the corresponding rankings judged by human assessors, while (b) and (e) show the correlation of the difficulty estimates and that of the subjectivity estimates respectively with the corresponding levels to which the images have been categorized by the assessors. Finally, (c) shows the images as points with coordinates being the difficulty and the subjectivity estimates, and has highlighted some images with noteworthy coordinates, while (d) shows these images.}
\label{figure:coherence}
\end{figure*}

The results of the sensitivity analysis described in section \ref{section:sensitivity_analysis} are shown in Tables \ref{table:robust_analysis} and \ref{table:robust_analysis1}. We can see from Table \ref{table:robust_analysis} that the average prediction accuracy of the SDR model with 1 latent preference is constantly higher than that of the model with 2 preferences over all the objective datasets. According to section \ref{section:sensitivity_analysis}, this result indicates there is just one underlying group of workers for each of the tasks who perceive the questions' correct answers in the same way. It also shows that even though the expertise-difficulty corruption introduced noises to the objective truths to form the actual responses, our model was still able to recover the number of underlying group of workers to be 1. From Table \ref{table:robust_analysis1}, our model with 2 preferences clearly outperforms itself with 1 preference across all the partially subjective tasks. This means multiple groups of workers have emerged due to the sufficient subjectivity of these tasks. Moreover, the table shows that further increasing the number of latent preferences to 3 no longer improves the performance. This has most likely been caused by over-fitting, and also suggests a two-dimensional latent space is accurate enough to explain the worker clustering effects emerged from these tasks. To further prove our model with 2 preferences can uncover the underlying groups of workers who have perceived the partially subjective tasks differently, we show the density of the workers' latent preference probabilities $\hat{\boldsymbol{\phi}}_i$ estimated by our model from the image data \cite{tian2012learning}. Due to a space limit, we only show two of them in Figure \ref{figure:density_plot}. According to \cite{tian2012learning}, the sub-task of judging whether images are beautiful is more subjective than the sub-task of identifying skies in images. This is re-confirmed by our model with its number of worker clusters for the former sub-task greater than that for the latter shown by Figures \ref{figure:density_plot_sky} and \ref{figure:density_plot_beauty}.

\begin{table}[t]
\centering
\scriptsize
\caption{ Average accuracy of our model with 1 and 2 latent preferences on predicting the held-out validation response of each worker over 4 objective tasks. \vspace{-.5cm}}
\begin{tabular}{|c|c|c|}
\hline
\multirow{ 2}{*}{\bf Dataset}&\multicolumn{2}{c|}{ {\bf The SDR model}}\\
&\bf m = 1&\bf m = 2\\
\hline
\bf Time & \bf0.8967 & 0.8915 \\ 
\bf Dog &  \bf0.6970 & 0.6625 \\
\bf Duck & \bf0.8427 &  0.8388 \\
\bf Product &  \bf 0.8396 &  0.8291\\
\hline
\end{tabular}
\label{table:robust_analysis}
\end{table}

\begin{table}[t]
\centering
\scriptsize
\caption{Average accuracy of our model with 1, 2 and 3 latent preferences on predicting the held-out validation response of each worker over 10 partially subjective tasks the first 5 of which are sub-tasks of the \textit{Image} task in \cite{tian2012learning}.\vspace{-.5cm}}
\begin{tabular}{|c|c|c|c|}
\hline
\multirow{ 2}{*}{\bf Dataset}&\multicolumn{3}{c|}{ {\bf The SDR model}}\\
&\bf m = 1&\bf m = 2&\bf m = 3\\
\hline
\bf Beauty 1 & 0.6736 & \bf0.6944& 0.6924\\ 
\bf Beauty 2 &  0.6914& \bf0.6998& 0.6937\\
\bf Sky & 0.8889& \bf0.8962 & 0.8862 \\
\bf Building &  0.8997&  \bf0.9026 & 0.9007\\
\bf Computer & 0.8098 & \bf0.8117 & 0.8074 \\
\bf Rel1 &  0.3956 &  \bf0.3985 & 0.3983 \\
\bf Rel2 &  0.4426 &  \bf0.4481 & 0.4481 \\
\bf Fashion & 0.7517 & \bf0.7589 & 0.7522 \\
\bf Face & 0.7181 & \bf0.7203 & 0.7123\\
\bf Adult &  0.7469 & \bf 0.7494 &0.7446\\
\hline
\end{tabular}
\label{table:robust_analysis1}
\end{table}

The results of the question correct answer prediction described in section \ref{section:true_label_pred} are listed in Table \ref{table:true_label_pred}. Across all the partially subjective datasets except the image data, the SDR model, based on the largest-group strategy for choosing the best worker clusters, is superior than the other 5 baselines\footnote{The performance of all the models on the image data (in Table \ref{table:data_summary}) has been too close to bear any useful information as for which of them is better since the number of the questions (i.e. 60) in the data is too small.}. Especially, for the tasks of relevance judgement 1\&2 and fashion judgement, our model is able to outperform the best baselines by 3\%, 1.5\% and 0.3\% with almost 54, 9 and 13 more correctly predicted question answers respectively. 
Since our model is reduced to being very similar to GLAD when dealing with the objective datasets, it has achieved very similar results as GLAD did in correct answer prediction over all the objective datasets except for the Duck data \cite{welinder2010multidimensional}. In this task, our model is superior than GLAD (0.69 versus 0.62 from GLAD). This suggests that our model is at least as robust as GLAD when predicting correct answers for objective tasks.

\begin{table}[t]
\centering
\scriptsize
\caption{ Accuracy of all the models on predicting the true answers of the four partially subjective datasets (the results for the \textit{Image} task are not included as the number of items in this task is too small to show any significant difference in the performance of different models). \vspace{-.5cm}}
\begin{tabular}{|c|c|c|c|c|c|c|}
\hline
\multirow{ 2}{*}{\bf Dataset}&\multicolumn{6}{c|}{ {\bf Question correct answer prediction}}\\
&\bf SDR&\bf MV &\bf GLAD&\bf DS&\bf CDS&\bf MdWC\\
\hline
\bf Rel1& \bf0.4998 &  0.4522 &  0.4457 & 0.4309 & 0.4697 & 0.4674 \\
\bf Rel2& \bf0.4752 & 0.4544 & 0.4567 & 0.4512 &0.4604 &  0.4586\\
\bf Fashion & \bf0.8733  &  0.8580 & 0.8689  & 0.8415 & 0.8463 & 0.8700\\
\bf Face & \bf0.6423 &  0.6404 & 0.6130  & 0.5924 & 0.5986 & 0.6079\\
\bf Adult & \bf0.7598 & 0.7568  &  0.7587 & 0.7534 & 0.7582 & 0.7556\\
\hline
\end{tabular}
\label{table:true_label_pred}
\end{table}

\begin{figure}[!t]
 	\centering
 	\includegraphics[width=3in]{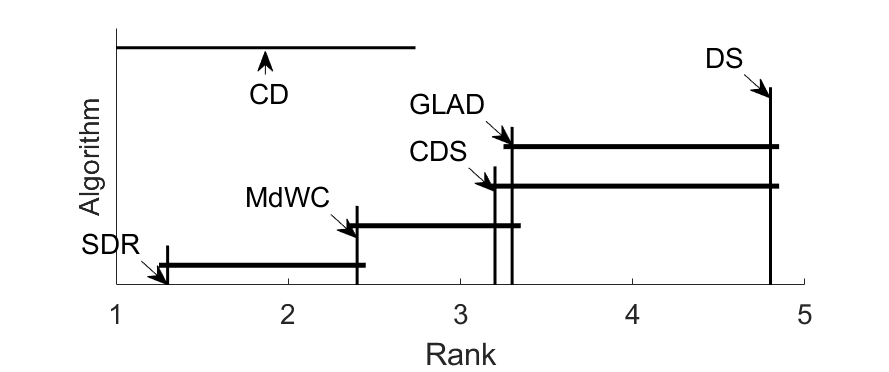}
 	\caption{Critical difference (CD) diagram of the Nemenyi post-hoc test ($\alpha=0.10$). The performance difference between two algorithms is significant if the gap between their ranks is larger than CD. There is a horizontal line connecting the two algorithms if the rank gap between them is smaller than CD.}
 	\label{rank}
 \end{figure}

The results of the worker answer prediction described in section \ref{section:worker_ans_pred} are shown in Table \ref{table:worker_label_pred}. We can see that our model is not the best on 3 out of the 10 partially subjective datasets, topped by different baselines. Despite that, our model has still performed adequately well (being second best on those datasets). We conjecture that this is because all these 3 datasets are with binary answer options which intrinsically constrain the answering behaviour of crowd-workers. This results in overall weaker correlations both in the worker responses and in the underlying correct answers across the questions. For the other 7 datasets, 5 of them are with more than two answer options, thus containing stronger answer correlations for our model to exploit to achieve better performance. 
To examine whether the difference in the worker answer prediction accuracy between any two algorithms is significant, we conducted the Nemenyi post-hoc test \cite{demvsar2006statistical} based on Table \ref{table:worker_label_pred}. The result is shown in Figure \ref{rank}, according to which the performance difference between SDR and either CDS, GLAD or DS is beyond the critical difference (CD), thereby being statistically significant. 

\begin{table}[t]
\centering
\scriptsize
\caption{Average Accuracy of all the models on predicting the unseen held-out test response of each worker across all the partially subjective datasets.\vspace{-.5cm}}
\label{table:worker_label_pred}
\begin{tabular}{|c|c|c|c|c|c|}
\hline
\multirow{ 2}{*}{\bf Dataset}&\multicolumn{5}{c|}{ {\bf Unseen worker answer prediction}}\\
&\bf SDR&\bf GLAD&\bf DS&\bf CDS&\bf MdWC\\
\hline
\bf Beauty 1 &  \bf 0.6974 &  0.6884 &  0.6256&  0.6927 & 0.6912\\ 
\bf Beauty 2 & 0.7006 &  \bf 0.7011  &  0.6796 &  0.6842 &  0.6998 \\
\bf Sky &  \bf 0.9014 &  0.8772 &  0.8801 &  0.8862 & 0.8903\\
\bf Building &  0.8987  &  0.8912 & 0.8956 & \bf 0.9006 &  0.8976\\
\bf Computer &  0.8284  & 0.8139 & 0.8115 &  0.8196 &  \bf 0.8336 \\
\bf Rel1 &  \bf 0.4067 & 0.4035 & 0.3654 & 0.3972 &  0.3987\\
\bf Rel2 & \bf 0.4386  & 0.4312 & 0.4257 & 0.4304 & 0.4340 \\
\bf Fashion & \bf 0.7659 & 0.7593 & 0.6977 & 0.7621 & 0.7633\\
\bf Face & \bf 0.7224 & 0.7193 & 0.6625 & 0.7081 & 0.7148\\
\bf Adult & \bf  0.7386 & 0.7347 & 0.6767 & 0.7312 & 0.7354\\
\hline
\end{tabular}
\end{table}

The results of the subjectivity and difficulty coherence evaluation have been summarised in Figure \ref{figure:coherence} which consists of 6 sub-figures. Figures \ref{figure:a} and \ref{figure:d} show overall there is a strong negative correlation between the model estimates and the rankings judged by human assessors. 
More specifically, the larger the estimate for either the difficulty or the subjectivity of an image, the higher it tends to be ranked by human assessors. Moreover, Figures \ref{figure:b} and \ref{figure:e} show that there exist clear positive correlations between the levels of difficulty and subjectivity into which the images get categorised by the human assessors, and the estimated values of these two properties inferred by the SDR model.

To support our argument about the efficacy of the SDR model in revealing the two key properties of images, we have selected four images highlighted in different colours in Figure \ref{figure:c} with their image ids. We can see that image 34 is inferred by our model to be both easy and objective as both of its estimates shown in Figure \ref{figure:c} are the smallest. This can be re-confirmed by visual inspection of the image in Figure \ref{figure:f}. It is very easy and clear to see that there is no sky in the image 34. Image 29 has been identified by our model to be hard with low subjectivity according to its estimates shown in Figure \ref{figure:c}. This is reasonable as the image indeed contains an extraterrestrial sky which is hard for novice workers to realise, while expert workers are able to realise and find the image objective. Images 2 and 23 both belong to the image beauty judgement task from \cite{tian2012learning} which requires workers to select 6 most beautiful images from 12 images. Our model has identified that image 2 is more subjective and harder to judge. This is probably because image 2 delivers a view of scenery which is more likely to resonate with workers while image 23 is merely showing an object. As a result, workers tend to show more different feelings and opinions towards image 2. On the other hand, image 23 does have better image quality and thus is easier for workers to make their decisions on whether it is beautiful or not.

\section{Conclusions}


In this paper, we have proposed the SDR (Subjectivity-and-Difficulty Response) model, a novel quality-control framework for crowdsourcing that is able to 
distinguish question subjectivity, which causes worker-specific truth for individual questions, from question difficulty, which determines the probability that a worker's response to each question equals her perceived subjective truth. Experiment results show that our model improves both the correct answer prediction for questions and the held-out unseen response prediction for crowd-workers compared to five baselines across numerous partially subjective crowdsourcing datasets. Moreover, our model shows robustness to both the objective and the partially subjective datasets by discovering the right numbers of underlying worker groups for them. Finally, our model is able to provide estimates of the difficulty and the subjectivity of questions that are consistent with the judgements from human assessors.

\bibliographystyle{ACM-Reference-Format}
\bibliography{sample-bibliography}

\end{document}